# p-DLA: A Predictive System Model for Onshore Oil and Gas Pipeline Dataset Classification and Monitoring - Part 1

E.N. Osegi

Department of Information and Communication Technology

National Open University of Nigeria

Lagos State, Nigeria.

System Analytics Laboratories (SAL)

Sure-GP Ltd

Port-Harcourt

Rivers State, Nigeria.

E-mail: nd.osegi@sure-gp.com

## Abstract

With the rise in militant activity and rogue behaviour in oil and gas regions around the world, oil pipeline disturbances is on the increase leading to huge losses to multinational operators and the countries where such facilities exist. However, this situation can be averted if adequate predictive monitoring schemes are put in place. We propose in the first part of this paper, an artificial intelligence predictive monitoring system capable of predictive classification and pattern recognition of pipeline datasets. The predictive system is based on a highly sparse predictive Deviant Learning Algorithm (p-DLA) designed to synthesize a sequence of memory predictive clusters for eventual monitoring, control and decision making. The DLA (p-DLA) is compared with a popular machine learning algorithm, the Long Short-Term Memory (LSTM) which is based on a temporal version of the standard feed-forward back-propagation trained artificial neural networks (ANNs). The results of simulations study show impressive results and validates the sparse memory predictive approach which favours the sub-synthesis of a highly compressed and low dimensional knowledge discovery and information prediction scheme. It also shows that the proposed new approach is competitive with a well-known and proven AI approach such as the LSTM.

## 1. Introduction

The continual militancy and vandalism that occur in oil and gas installations have led to high levels of insecurity to industry operators disrupting their operations and as a consequence, leading to huge losses to oil industry operators. In line with the need to prevent/or reduce further disruptions to oil and gas operations, worldwide research in the area of oil/gas pipeline facility protection have resulted in the development of many proposals on how to tackle the oil and gas insurgency, in particular, in the Niger Delta region of Nigeria – the oil rich hub of West Africa. Popular among the list of proposed solutions is the use of real time monitoring systems based on several existing and future technologies. For instance, the use of overhead and underwater surveillance have been briefly surveyed and wireless monitoring systems proposed in [1] and [2] respectively. This has the advantage of real time visualization and transmission but also adds with it the complexity of processing vast amount of video or image data. Another interesting area is the use of artificial experts and robotics for smart monitoring and sensing in oil fields. See for instance the review in Shukla and Karki [3, 4]. For instance, in [5, 6] and in [7], the expert systems approach and robotics have been deployed for onshore pipeline analysis and underwater pipeline monitoring respectively. While some systems use wired means of detection, the benefits of wireless monitoring systems cannot be overemphasized [8].

However, little work has been done in the area of predictive classification of pipeline data, particularly as it pertains the development and analysis of real world pipeline datasets. As have been shown in [9, 10], and in [11], predictive systems using highly sparse cortical learning algorithms can prove useful in the detection of likely anomalies/or defects particularly in an online and unsupervised data monitoring system.

In this paper, we propose the use of a predictive memory system based on the Deviant Learning Algorithm (DLA); the memory is obtained using the sparse generative cortical learning algorithm designed for stream data processing [12]. Our primary purpose in the current version of this paper is to validate the effectiveness of this new AI technique for predictive classification of two pipeline datasets that have been developed by other researchers in [13] and [14]. These datasets are chosen due to their interpretability and conformance to real world observations in oil and gas environments and will set the course of direction in subsequent versions of this paper.

## 2. Concept of Predictive Monitoring and Cortical-like Sparse Memory Generation

The idea of predictive monitoring using sparse-generative cortical learning approaches is not entirely new to the machine learning community. In [15] the idea of memory predictive system was founded which engineered the development of variations of memory predictive architectures referred to now as Hierarchical Temporal Memory; these architectures have been developed in [16], [17], [18] which is based on the cortical learning algorithms, and in [19] which is based on the Bayesian Belief Network (BBN). Sparse coding originally discovered and formally presented in [20] provided the necessary foundation for HTM cortical learning algorithms herein called

HTM-CLA. While sparse coding presented the necessary foundations for cortical-like algorithms, predictive coding takes it a step further. As described in Huang and Rao [21], predictive coding presents a unifying paradigm for explaining the functional properties of key neural tissue by actively predicting hidden causes of incoming sensory information; this forms the basis of most modern cortical-like predictive monitoring systems such as that proposed in this paper.

## 2.1 Skip-Sequence iterator for sparse-predictive memory systems

In this part of this paper, we introduce the concept of skip-sequence iterator for cost-effective generation of highly sparse memories. Skip-sequences is basically a form of drop-out where a portion of the input observation is sequentially skipped. This concept is illustrated pictorially in Figure 1. The skip-sequence approach creates an affordable, faster and hence less power-hungry mining process during data predictive learning which can help in reducing the dimensionality in data and hence avoid the "curse of dimensionality" in the DLA's operation. Reducing the dimensionality in data has been proven to be a useful and very important task particularly when the datasets grow in size and memory needs to be conserved [22]. However, as will be seen later in section 4, this comes with the price of lower predictive classification accuracies and recognition abilities.

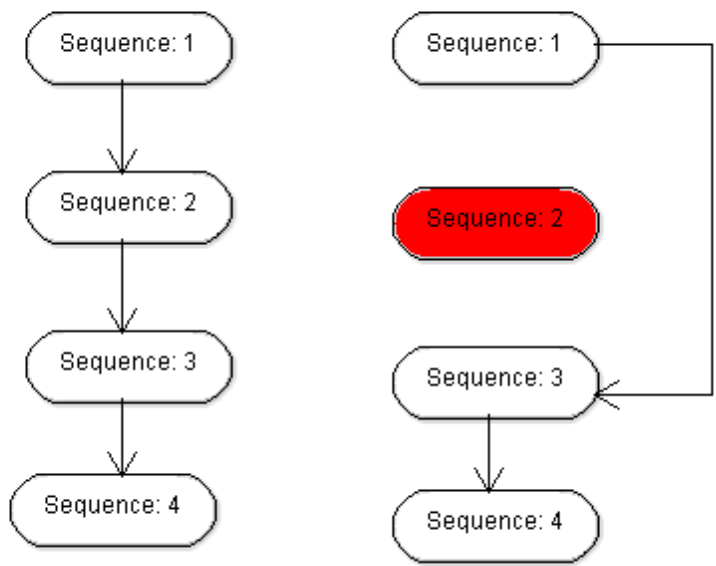

Fig1. Illustration of two activities of Skip-Sequences; note that sequence 2 is skipped in the second activity leading to a sparse memory sequence-of-sequences

## 2.2 Learning Extent Theory (LET)

The learning extent theory is in line with the birth-death principle and is based on the hypothesis that learning performance increases as learning units increase. Here, birth refers to the instantiation of a learning machine and the start of its learning operations. The primary units of learning in the DLA is the "symbolic integer". The integers along with a DLA processor [12] forms a sparse field of memory patterns. At birth only a few learning units (symbolic integers) is available during learning; but as the learning machine ages, the learning units increases and the machine starts to learn more complex tasks i.e. its learning performance appreciates due to a higher integer capture. However, in line with the death principle, at a certain learning extent (threshold) the machine should start to degrade in performance with reducing learning units until the machine dies off i.e. performance degrades to a point where learning is no longer effective. This process of variational learning can also be interpreted in terms of the cortical synapses [16] where the integer learning units is replaced with the cortical synapses.

## 3. Systems Architectural Modelling

The proposed systems architecture for predictive classification of pipeline datasets is shown in Figure 2. It is based on the DLA developed in [12] and consists of the following key units:
- The Pipeline data unit which serves as container for pipeline dataset
- An Integer Encoder for transforming the pipeline input dataset into a mixed-integer sparse distributed representation (SDR)
- A Sequential Pre-prediction unit for initial mixed-integer pre-processing of the incoming sensory input. This generates a sequence of SDRs in the memory space.
- A Sequential Post-prediction unit that performs further pre-processing of the Sequential Pre-prediction
- An optional Backward Additive Deviant Computing Unit that allows extrapolations to be made on the DLA's post-predictive memories
- A Temporal classifier for evaluating the predictive classification accuracy of the entire system. The classification accuracy is computed as [12]:

$$MAPCA = \left( \frac{\sum_{i}^{n}(|y_{(i)} - \hat{y}_{(i)}| < tol)}{n_z} \right) * 100 \qquad (1)$$

where,

    MAPCA = mean absolute percentage classification accuracy

    $y$ = the observed data exemplars

$\hat{y}$ = the model's predictions of $y$

$n_z$ = size of the observation matrix, and

*tol* = a tolerance constraint.

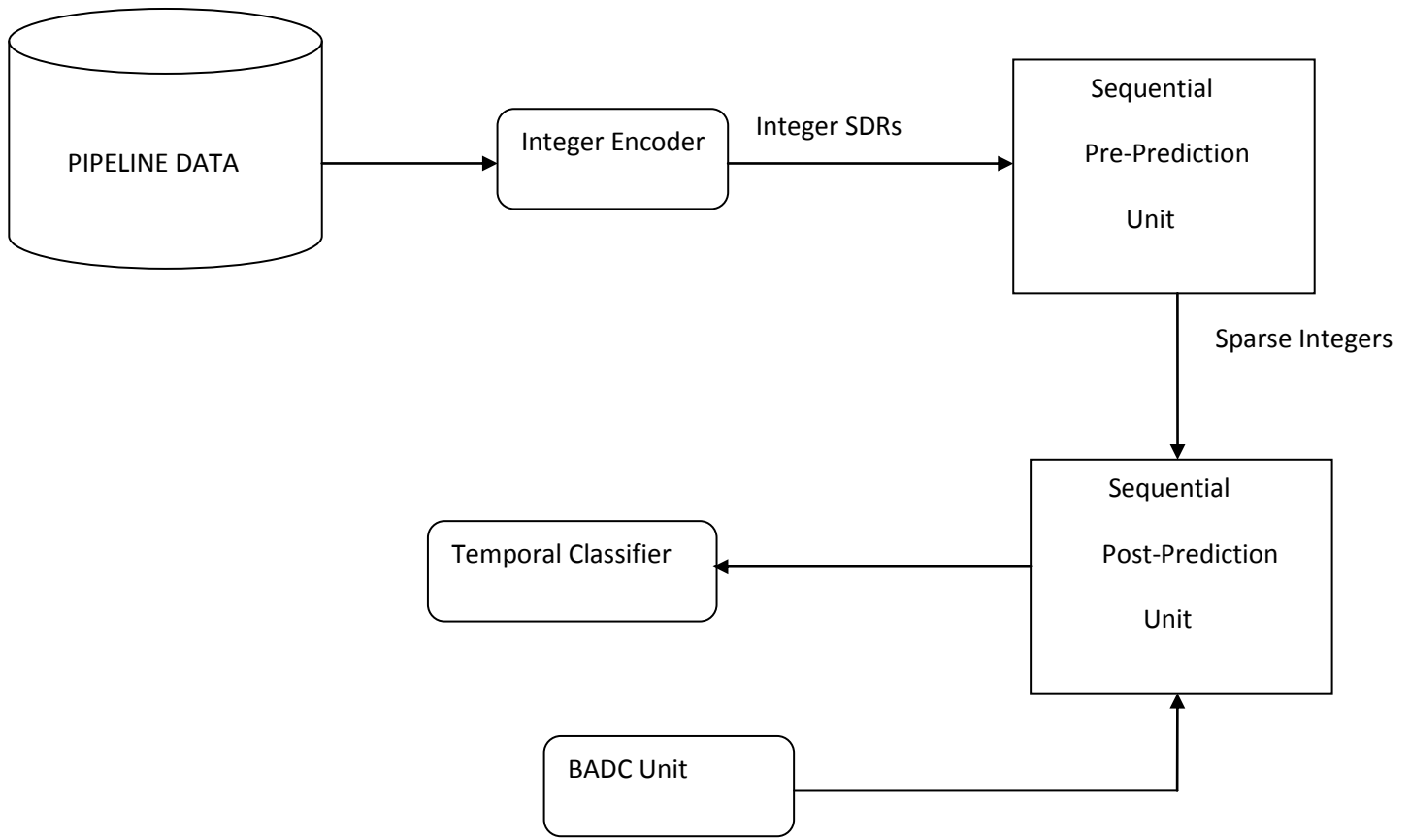

Fig2. Proposed Systems Modelling Architecture

### 3.1 Sequential Predictions

Sequential predictions follows a two-stage processing regime. First, the sparse set of input data is first captured into a memory store. Then, the sparse memory is processed dynamically online using the DLA in a temporal manner. In order to perform extrapolations, the DLA uses the backward additive deviant computing formula to compute an aggregated expression as [12]:

$$K_{avg}^t = \frac{\sum_{j=1}^{n}|K_n^t - K_{seq}^t|}{n} \quad \{K_{seq}^t = K_j^t, \quad j = 1,2,..n-1, j \in Z^+ \quad (2)$$

where,

$K_n^t$ = nth memorized sequence chunk at time, t

$K_{seq}^t$ = memorized sequence chunks at time steps of t

$n$ = number of previously memorized sequence chunks

Using (2), the DLA's numeric prediction can then be computed as:

$$K_p^t = K_{avg}^t, + K_n^t \quad (3)$$

### 4. Experimental Results and Discussions

Simulation experiments have been performed using the pipeline feature datasets developed in [13] and [14]. The dataset used in [13] is a small feature set with less than 20 exemplars while the dataset used in [14] is a much larger one with about 190 exemplars.

The experiments have been divided into three parts: for the first part, comparative pattern recognition simulations with the LSTM algorithm is performed as proof of correctness test using the dataset in [13] with fixed values for the Variational Learning Extent (VLE) and the skip-sequence (SKS) values; for the second part, we study the operational effect of VLE for the dataset in [14] while keeping the SKS value constant at 1 unit. The third and final part examines the effect of static SKS tuning on the DLA predictions with the VLE set to its maximum value using the dataset in [14].

### 4.1 Experiment 1

For the first set of experiments, we perform comparative simulations with a popular machine learning approach – the Long Short-Term Memory (LSTM) using the dataset in [13]. The predicted values from the DLA and LSTM simulations are as shown in Figure 3. The best and most stable parameters are used for both algorithms and are given in Appendix A.

```
DLA:

Vehicle passing 2.02 49.78 40.75 2.42 4.57 0.27 0.08 0.03 0.06 20

Machine excavation 0.31 46.78 48.39 1.58 2.45 0.35 0.06 0.02 0.03 34

Machine excavation 0.17 40.01 55.23 1.27 2.73 0.52 0.05 0.01 0.01 36

Manual digging 25.97 0.81 9.37 39.77 6.61 6.21 7.91 1.85 1.47 22

LSTM:

Manual digging 23.31 0.85 10.01 41.86 1.97 0.91 0.03 0.03 0.03 0.03 1.93 0.05 0.02 0.03 0.03 0.03 1.9
```

Fig3. The DLA and LSTM predictions

### 4.2 Experiment 2

In this experiment, we study the influence of increasing SKS values on the classification accuracy and sparse prediction representation of the DLA using the dataset in [14]. Using the SKS concept can further increase the sparsity of the DLA's prediction and also reduce the time taken in processing huge datasets. For each simulation run, the SKS value is changed in increments of 5units. The simulation (trend) plots of integer-prediction data is as shown in Figure 4 while the classification accuracies at the different SKS values is given in Table 1.

TABLE 1: Classification accuracies for the DLA at different SKS values

|  | SKS – 1 unit | SKS – 5 units | SKS – 10 units |
|---|---|---|---|
| **Percent accuracy (%)** | 99.9316 | 88.1605 | 87.4447 |

### 4.3 Experiment 3

In this experiment, we study the influence of increasing VLE on the classification accuracy and sparse prediction representation of the DLA using the dataset in [14]. For each simulation run, the learning extent is changed from an initial value of 60 units to a final value of 100 units. The choice of values is just sufficient to demonstrate the influence of the VLE on the DLA's predictions. The simulation plots capturing this effect is as shown in Figure 5 while the classification accuracies for the different learning extents is given in Table 2.

TABLE 2: Classification accuracies for the DLA at different learning extents (l_ext)

|  | $l_{ext}$ – 60 units | $l_{ext}$ – 80 units | $l_{ext}$ – 100 units |
|---|---|---|---|
| **Percent accuracy (%)** | 99.9316 | 88.1605 | 87.4447 |

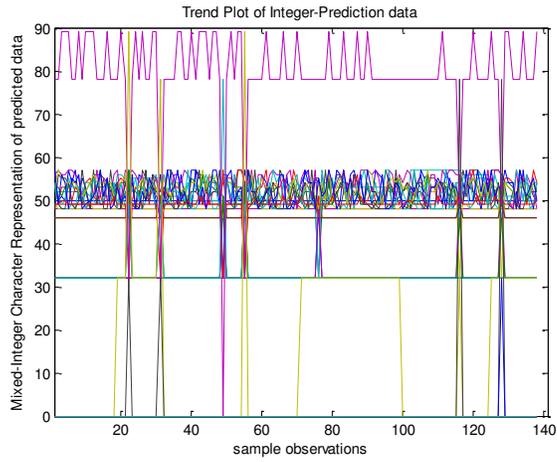

(a) Trend plot at SKS value of 1 unit

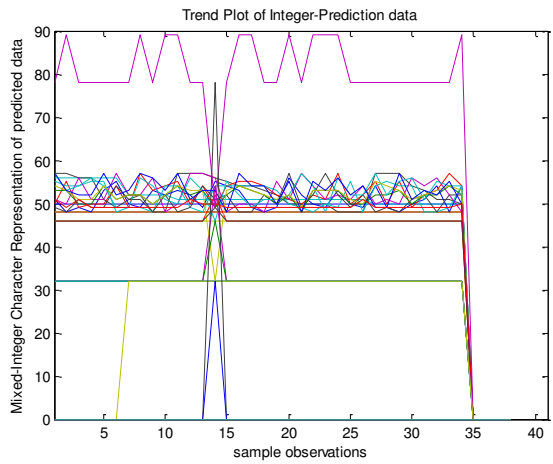

(b) Trend plot at SKS value of 5 units

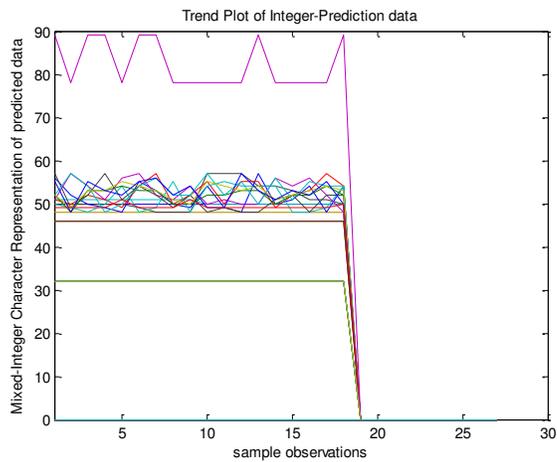

(c) Trend plot at SKS value of 10 units

Fig.4. Visualization of trend plots for different SKS values using the DLA program

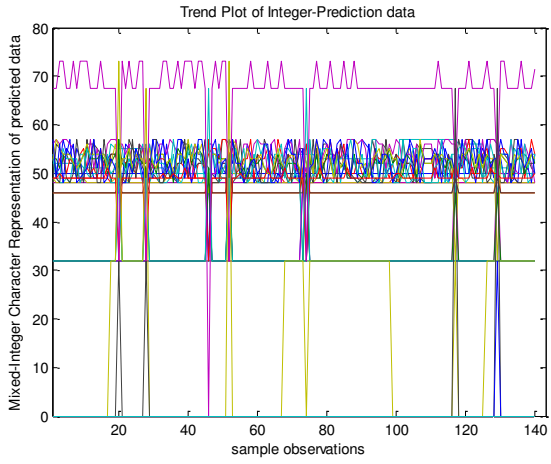

(a) Trend plot at learning extent of 60 units

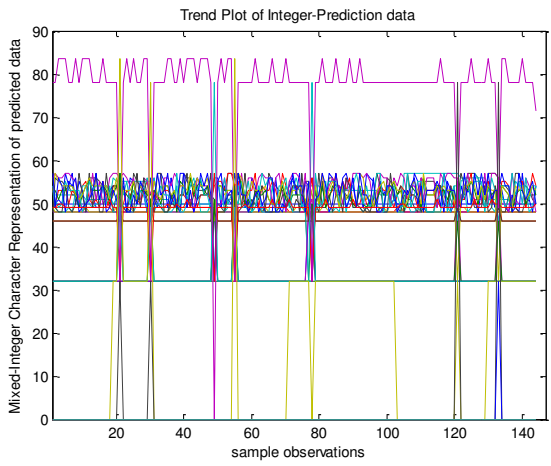

(b) Trend plot at learning extent of 80 units

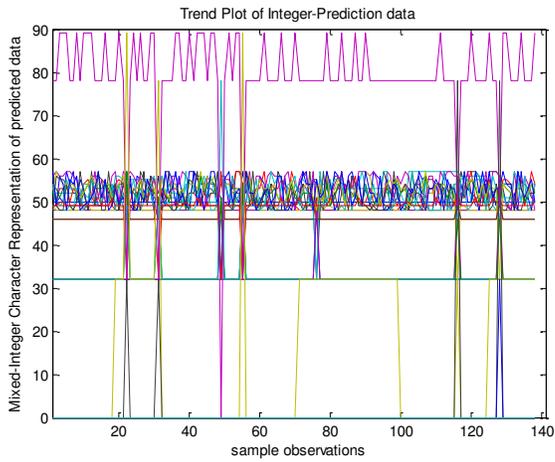

(c) Trend plot at learning extent of 100 units

Fig.5. VLE trend plots using the DLA program

### 4.4 Discussions

In Experiment 1, the DLA recognition shows higher multiple predictions than the LSTM (see Figure 2). This shows that the DLA can learn more units and is more likely to give a higher recognition rate than the LSTM for this dataset. From the trend (visualization) plots in experiment 2 (see Figure 4), it is obvious that increasing SKS values leads to a highly sparse representation. However, as shown in Table 1, the classification accuracy degrades with increasing SKS values. Thus, the reduction procedure due to incremental SKSs must be handled with care in highly critical monitoring operations.

The results shown in Experiment 3 confirms the learning extent theory (LET). With increasing learning extents, we should expect a better performance as illustrated in the trend plots shown in Figure 5(a) to 5(c). The higher the learning extent the sharper the peak pulses in the integer-prediction data. The classification accuracies are given in Table 2 and shows that the accuracy will improve with higher learning extent. However, there is a limit on the maximum possible accuracy, and further increases on the learning extent have no significant influence over the DLA's prediction.

### 5. Conclusion

The abilities of a novel machine learning (ML) algorithm – the Deviant Learning Algorithm (DLA), in performing predictive classification of onshore pipeline incidence/threat datasets have been demonstrated in this research paper. The model has been shown to be comparable in memory prediction abilities with a proven ML algorithm – the Long Short-Term Memory (LSTM). Interestingly, the DLA was capable of multiple predictions of a pipeline dataset whereas the LSTM was not able to achieve this important operation. The DLA also shows promising accuracies of greater than 85% with higher sparsity. This has the advantage of lower cost in sequential memory processing when performing predictive monitoring operations.

In part 2 of this paper, we will focus on a more efficient methodology for a real-time predictive pipeline monitoring system with the hope that this will yield an improved set of results which will in turn unlock the potentials of sparse memory predictive systems for real-time monitoring tasks.

### Acknowledgement

This study received no funding from any source or agency. Source codes for the DLA simulations and dataset are available at the Matlab central website:

www.matlabcentral.com

# APPENDIX A: PARAMETERS FOR THE LSTM AND THE DLA

**Table A.1:** LSTM PARAMETERS

| Parameter | min | Max |
|---|---|---|
| Hidden sizes | 20 | 20 |
| Character size | 5 | 5 |
| Learning Rate | 0.0 | 0.01 |
| L2 Regularization Strength | 0.000001 | 0.000001 |
| Clip Value | 0.000005 | 0.05 |
| Softmax sample temperature | 0.0 | 0.1 |

**Table A.2: DLA PARAMETERS**

| Parameter | min | max |
|---|---|---|
| Learning extent | 121 | 125 |
| Time limit | 10 | 10 |
| Initial permanence value | 0 | 0 |
| Store threshold | 120 | 120 |
| Tolerance constraint | 0.05 | 0.05 |